\documentclass[lettersize,journal]{IEEEtran}
\usepackage{amsmath,amsfonts}
\usepackage{algorithmic}
\usepackage{algorithm}
\usepackage{array}
\usepackage[caption=false,font=normalsize,labelfont=sf,textfont=sf]{subfig}
\usepackage{textcomp}
\usepackage{stfloats}
\usepackage{url}
\usepackage{verbatim}
\usepackage{graphicx}
\usepackage{cite}
\usepackage{multirow}
\usepackage{cleveref}
\usepackage{eccvabbrv}
\usepackage{booktabs}

\newcommand{\supp}{\textit{Supp. Mat.}\xspace}

\begin{document}

\title{TalkCLIP: Talking Head Generation with Text-Guided Expressive Speaking Styles}

\author{Yifeng Ma\IEEEauthorrefmark{1}, Suzhen Wang\IEEEauthorrefmark{1},  Yu Ding, Bowen Ma, \\ Tangjie Lv, Changjie Fan, Zhipeng Hu,   Zhidong Deng, Xin Yu
        % <-this % stops a space
\thanks{\IEEEauthorrefmark{1} Equal contribution.}
% \thanks{\IEEEauthorrefmark{2} Corresponding author.}
\thanks{Yifeng Ma and Zhidong Deng are with Department of Computer Science and Technology, BNRist, THUAI, State Key Laboratory of Intelligent Technology and Systems, Tsinghua University, China. E-mail: \{mayf18@mails., michael@\}tsinghua.edu.cn.}
\thanks{Suzhen Wang, Yu Ding, Bowen Ma, Tanjie Lv, Changjie Fan and Zhipeng Hu are with Fuxi AI Lab, Netease, Hangzhou, Zhejiang, China. E-mail: \{wangsuzhen, dingyu01, mabowen, hzlvtangjie, fanchangjie, zphu\}@corp.netease.com.}
\thanks{Xin Yu is with the School of Computer Science, the University of Queensland, Brisbane, Australia. E-mail: xin.yu@uq.edu.au}}

\markboth{Journal of \LaTeX\ Class Files,~Vol.~14, No.~8, August~2021}%
{Shell \MakeLowercase{\textit{et al.}}: A Sample Article Using IEEEtran.cls for IEEE Journals}


\maketitle

\begin{figure*}
    \centering
    \includegraphics[width=\textwidth]{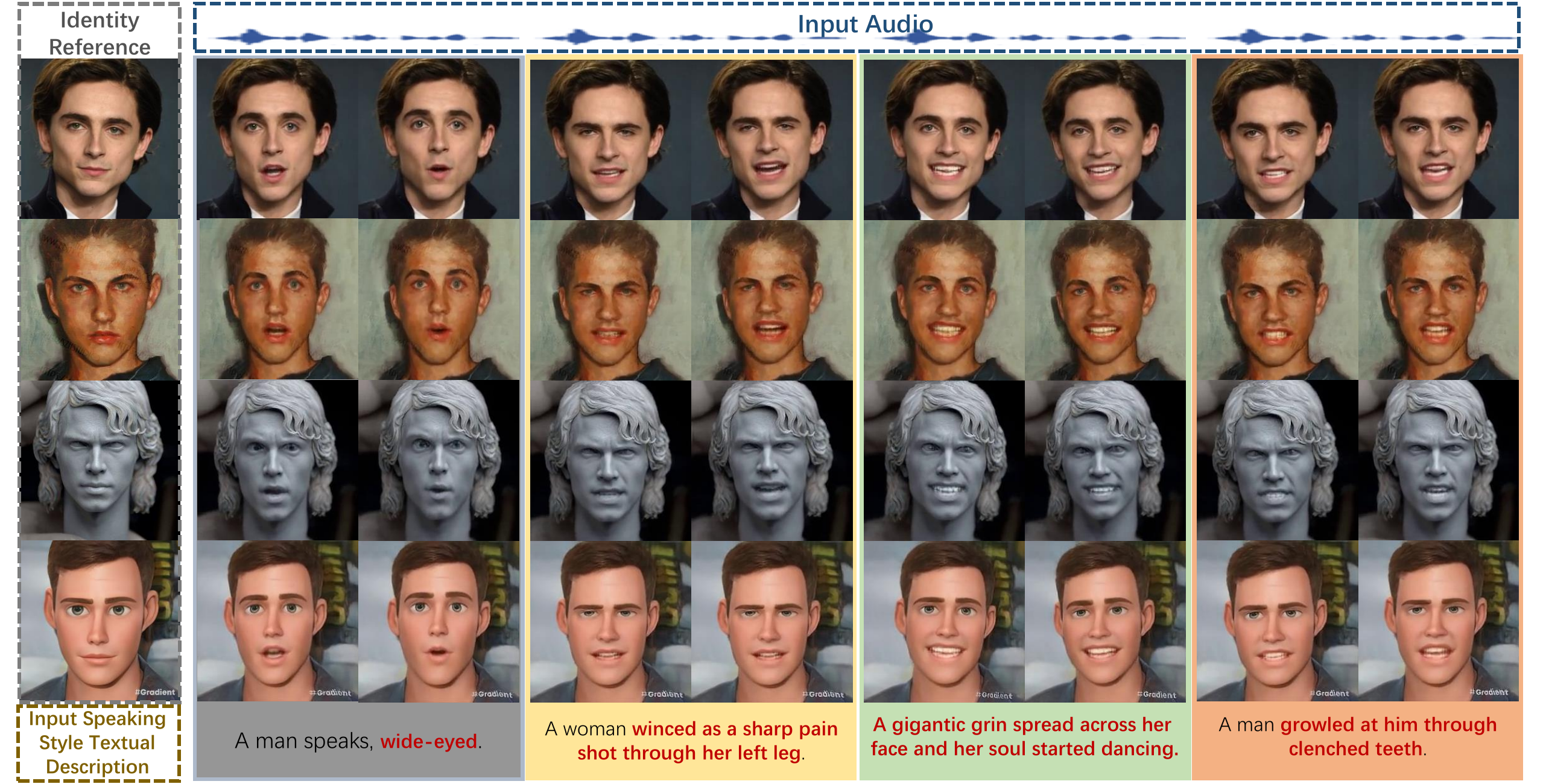}
      \caption{Results of TalkCLIP. Given natural language text that describes the desired speaking style, TalkCLIP can produce audio-driven talking head videos with the specified speaking style. The text can be unseen during training (\emph{Red texts} indicate \emph{unseen words} in training).}
       \label{fig:teaser}
\end{figure*}

\begin{abstract}
Audio-driven talking head generation has drawn growing attention. To produce talking head videos with desired facial expressions, previous methods rely on extra reference videos to provide expression information, which may be difficult to find and hence limits their usage. In this work, we propose TalkCLIP, a framework that can generate talking heads where the expressions are specified by natural language, hence allowing for specifying expressions more conveniently. To model the mapping from text to expressions, we first construct a text-video paired talking head dataset where each video has diverse text descriptions that depict both coarse-grained emotions and fine-grained facial movements. Leveraging the proposed dataset, we introduce a CLIP-based style encoder that projects natural language-based descriptions to the representations of expressions. TalkCLIP can even infer expressions for descriptions unseen during training. TalkCLIP can also use text to modulate expression intensity and edit expressions. Extensive experiments demonstrate that TalkCLIP achieves the advanced capability of generating photo-realistic talking heads with vivid facial expressions guided by text descriptions.
\end{abstract}


\section{Introduction}
\IEEEPARstart{I}{n} recent years, the audio-visual industry has experienced rapid growth, leading to the emergence of many cutting-edge techniques. As one of them, 
audio-driven talking head generation aims to generate a talking head video with the speaker specified by an input portrait and facial movements driven by an input audio clip. The task has gained increasing interest due to its diverse applications, including virtual avatars~\cite{10229247}, short video creation~\cite{yi2022predicting,wang2022one}, online courses, computer games, and virtual reality. To generate realistic talking heads, it is important to generate and control expressions, which are also called speaking styles~\cite{ma2023styletalk}. To control expressions, previous methods~\cite{yi2022predicting,liang2022expressive
,ma2023styletalk} mainly rely on extra reference videos to provide style information. However, finding a reference video with desired expressions may require much manual effort and hence is inconvenient. 

Instead of relying on videos, in this paper, we aim to use natural language text to specify the desired speaking style, an approach that is more convenient for real-world applications.
Learning the mapping from natural language text to speaking styles requires a talking head dataset containing text descriptions that depict speaking styles with diversity and in detail. However, existing datasets~\cite{wang2020mead,zhu2022celebvhq,yu2022celebvtext} only depict a limited number of emotions and face actions. To bridge the gap, we construct a talking head dataset, called TA-MEAD, that contains diverse and detailed text descriptions for speaking styles. The videos of TA-MEAD come from an existing emotional dataset MEAD~\cite{wang2020mead}. The text descriptions cover not only diverse coarse-level emotions, such as \emph{ecstatic}, but also fine-grained facial movements, such as \emph{wrinkle the nose}. The coarse-level emotion labels are manually labeled in a labor-efficient way, and the descriptions of fine-grained facial movements are automatically generated through AU detection.

Building upon TA-MEAD, we propose TalkCLIP, a framework that can generate audio-driven talking head videos with text-guided speaking styles. At the core of TalkCLIP is a text-to-speaking-style (T2SS) encoder that can map text descriptions to speaking styles. T2SS encoder uses CLIP text encoder as the backbone, leveraging CLIP's rich knowledge in natural language to achieve better generalization on real-world texts. Since CLIP is trained with text-image pairs that are static, while speaking styles involve facial movements that are dynamic, there is a gap between CLIP text embeddings and speaking style embeddings. To bridge this gap, we develop an adapter network within the T2SS encoder to adapt the CLIP text embeddings to the speaking style space. Furthermore, due to its cross-modal nature, text-to-speaking-style mapping is difficult to learn. To ease the learning difficulty, we propose a video-guided learning procedure where we first train a style encoder that extracts speaking styles from videos and then use it as a "teacher" to guide the learning of T2SS encoder.

Extensive experiments demonstrate that TA-MEAD's text descriptions can generally describe emotions accurately and TalkCLIP can generate talking head videos with the exact speaking styles specified by natural language. TalkCLIP can even generalize to out-of-domain text descriptions, including new words, phrases, and sentence structures not provided during training (\cref{fig:teaser}). TalkCLIP can also use text to modulate expression intensity and edit expressions. We hope that TA-MEAD and TalkCLIP can inspire research on text-to-face-motion generation.

\section{Related Work}

\noindent\textbf{Audio-Driven Talking Head Generation.}
Audio-driven methods can be divided into two categories: person-specific and person-agnostic methods. 

Person-specific methods~\cite{suwajanakorn2017synthesizing,ji2021audio,wang2020mead,1468149,865480} are limited to generating talking head videos for speakers who were present during the training phase. 
Most of these person-specific methods~\cite{thies2020neural, li2021write,ji2021audio, zhang20213d, zhang2021facial} initially create 3D facial animations and then utilize them to generate realistic talking head videos. Recently, some advanced techniques~\cite{guo2021ad,liu2022semantic}, such as SD-NeRF~\cite{10229247}, have utilized neural radiance fields to model talking heads, producing high-quality videos with a realistic appearance.

Person-agnostic methods aim to generate talking head videos for arbitrary speakers who have not been seen during training. Early methods~\cite{chung2017you,chen2018lip,zhou2019talking,chen2019hierarchical,prajwal2020lip,9917325} focused on achieving accurate lip synchronization. Subsequent methods~\cite{chen2020talking,zhou2020makelttalk,zhang2021flow,wang2021audio2head,zhou2021pose,wang2022one,yu2023talking,sadoughi2019speech,sinha2022emotion,yi2022predicting,10229247,10543093,vougioukas2019realistic}, including TFDCK~\cite{9681173} turned their attention to generating natural facial expressions and head poses. Recent methods~\cite{ji2022eamm,liang2022expressive,ma2023styletalk,wu2021imitating} have been proposed to mimic the speaking style of a reference video. Several methods~\cite{wang2022one,wang2021audio2head,ji2022eamm,ma2023styletalk,zhang2023sadtalker} first generate intermediate representations such as 3DMM coefficients and keypoints from audio, and rendering them to realistic talking videos. However, previous methods, such as Yi et al.~\cite{yi2022predicting} can only control speaking style using videos, which may be difficult to find. This work explores controlling speaking styles using text, which is more convenient.

\noindent\textbf{Text-Guided Visual Content Generation.}
There has been extensive research in generating visual content from text, including images, videos, and human motions. Text-to-image generation techniques~\cite{ramesh2021zero,saharia2022photorealistic,rombach2022high} have shown promising results. Some methods~\cite{xia2021towards,sun2022anyface} produce facial images where facial attributes such as face shape and hair color are controlled by the input text. In contrast, our approach utilizes text to guide the generation of facial expressions such as frowning, wrinkling the nose, and expressing joy.

Recently, text-to-video generation has garnered increasing attention from researchers~\cite{singer2022make,ho2022imagen,ho2022video,hong2022cogvideo}. Of these, text-driven motion generation\cite{tevet2022motionclip,ghosh2021synthesis}, aiming to generate human motions in line with text descriptions, is closely related to our work. 
% However, generating plausible talking head videos is particularly challenging due to the sensitivity of humans to facial artifacts.
In this work, we aim to address this challenge by precisely controlling the facial expressions of the talking head using text. 
%This will allow for more flexible control over the facial expressions of a talking head video.
There are some prior attempts in text-guided expression control in talking head generation.
However, Xu \etal~\cite{xu2023high} only supports short textual labels, while Gan \etal~\cite{gan2023efficient} requires additional training for each new textual prompt. In contrast, our method aims to achieve text-guided control using natural language prompts without the need for further training.

\section{TA-MEAD Dataset}\label{data_preparation}

\begin{figure*}[t]
  \centering
  \includegraphics[width=\linewidth]{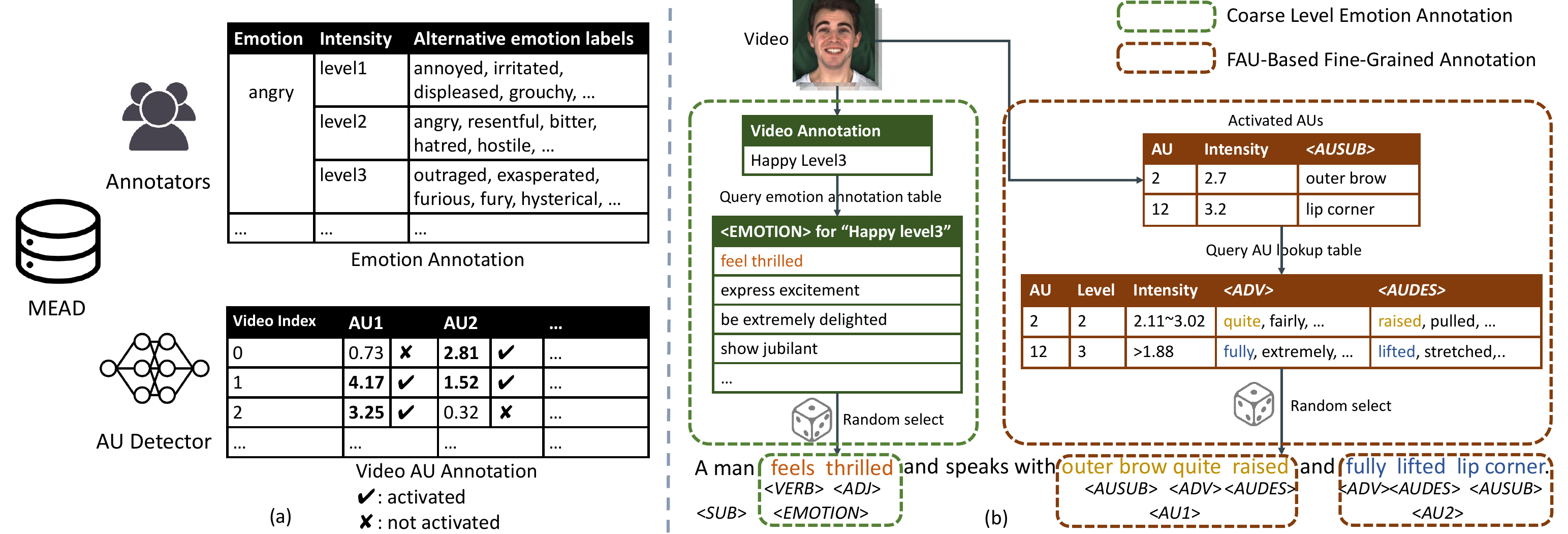}
  \caption{(a) The two types of video annotation. We recruit annotators to annotate the emotion of emotion-consistent video groups and obtain the emotion annotation table. We utilize an off-the-shelf AU detector to detect the AU intensity of each video and obtain the video AU annotation table. (b) The pipeline of automatically constructing the description sentence for the video.}
  \label{fig:annotation}
\end{figure*}

To model the mapping from natural language and speaking styles, we develop a talking head dataset that contains diverse natural language text annotations for facial expressions. The dataset is based on MEAD and called Text-Annotated MEAD (TA-MEAD). When describing expressions, people sometimes use emotion labels such as \emph{excited} or \emph{hysterical}, and at other times, they describe specific facial movements, such as \emph{frowning}, \emph{raising lip corners}. We call the former type coase-level emotion and the latter type fine-grained level expressions and annotate them in a labor-efficient way. After obtaining the annotations, we devise rules to generate diverse text descriptions for facial expressions. \cref{fig:annotation} shows the overview.

% To understand the semantic correlations between natural language and speaking styles, we created a dataset called Text-Annotated MEAD (TA-MEAD), consisting of text-annotated talking head videos. We used a labor-efficient approach to annotate the videos in two types: coarse-level emotion annotations and AU-based fine-grained annotations.

\noindent\textbf{Coarse-Level Emotion Annotation.}
We aim to annotate the diverse human expressions with as varied emotional labels as possible. 
The emotional talking head dataset currently available only includes a limited number of emotion labels (8 for the MEAD dataset), which is insufficient to depict the diverse range of human expressions. To address this issue, we adopt 135 emotion categories proposed in Chen \etal\cite{chen2022semantic} to annotate the videos, covering most distinctive emotional human expressions. Since existing emotion recognition models~\cite{chen2022semantic} fail to classify the 135 emotion categories correctly, we decide to annotate the emotion labels manually. Manually annotating thousands of videos is labor-intensive, but we find a labor-efficient way. We observe that in the MEAD dataset, the expressions in a group of videos from the same emotion category and the same emotion intensity are mainly consistent. Therefore, we can annotate a small portion of videos in a group and then the annotated labels can be shared by all videos in the group. MEAD consists of 22 groups, including one neutral emotion and seven other emotions with three levels of intensity. For each group, we sample three videos from each of the 48 speakers, resulting in a total of 144 videos for annotations. Five professional annotators are invited to collectively determine which emotions out of 135 can be used to annotate the samples in a group. The selected emotion labels constitute the coase-level emotion annotation for the group. \supp shows an example of coase-level emotion annotation.

\noindent\textbf{AU-based Fine-grained Annotation.}
To annotate fine-grained facial movements, we leverage rich facial information in the Facial Action Coding System (FACS)~\cite{ekman2002facial}. FACS breaks down facial expressions into individual muscle movements, known as Action Units (AUs)~\cite{ekman2002facial}. Each AU has a detailed text description that can easily be converted into text annotations. Therefore, we first use an off-the-shelf AU detector~\cite{ma2022facial} to detect AU and then use AU descriptions to construct annotation. Specifically,  for each video, we use the AU detector to obtain the AU intensity of each frame. The AU intensity of all frames is averaged to represent the AU intensity for the video. If the video AU intensity is higher than a predefined threshold, we think the AU is activated in the video. We also define thresholds to decide at which intensity level the AU is activated. This allows our method to model the intensity of face motions. 
% Finally, we convert activated AUs and their intensity levels into text annotations using text from FACS descriptions.

\noindent\textbf{Description Sentence Generation.}
% Once we obtain the emotion labels, activated AUs, and their intensities, we are able to generate text descriptions for each video.
Once we obtain the emotion labels, activated AUs, and their intensities, we are able to generate diverse text descriptions for each video. 
% The process of generating the descriptions is illustrated in Figure~\ref{fig:annotation}. 
The general format is  [\textit{$\langle$SUB$\rangle$ $\langle$EMOTION$\rangle$ and speaks with $\langle$AU$\rangle$}]. Here, \textit{$\langle$SUB$\rangle$} refers to the subject, while \textit{$\langle$EMOTION$\rangle$} and \textit{$\langle$AU$\rangle$} correspond to the coarse-level and fine-grained annotations, respectively.
 
To generate the text descriptions, we first randomly select an annotated emotion label from the alternative ones for \textit{$\langle$EMOTION$\rangle$}. 
As for \textit{$\langle$AU$\rangle$}, we follow the text generation approach proposed by Hong \etal~\cite{hong2020face}. This involves identifying the activated AUs (\emph{e.g.} eyebrows) and their corresponding descriptions (\emph{e.g.} lifted). 
For example, if the AU \emph{Outer Brow Raiser} is activated and the intensity is high, we annotate it as \emph{A man speaks with outer brow extremely lifted.}
Furthermore, during training, to improve generalization, we randomly drop the emotion or AU descriptions. Thanks to our detailed description of expression motion patterns, our subsequent methods can effectively learn the mapping from text to expression characteristics with a high degree of accuracy. Our TA-MEAD dataset will be released.

\section{TalkCLIP Framework}\label{talkclipframework}
%--Inspired by recent works that effectively learn visual concepts from natural language supervision, we utilize the extensive semantic knowledge embedded in CLIP, which has achieved striking performance when adapted to a wide range of downstream tasks.

% Based on TA-MEAD, we propose a novel framework for generating talking head videos with text-guided speaking styles. 

\begin{figure*}[t]
  \centering
  \includegraphics[width=0.98\textwidth]{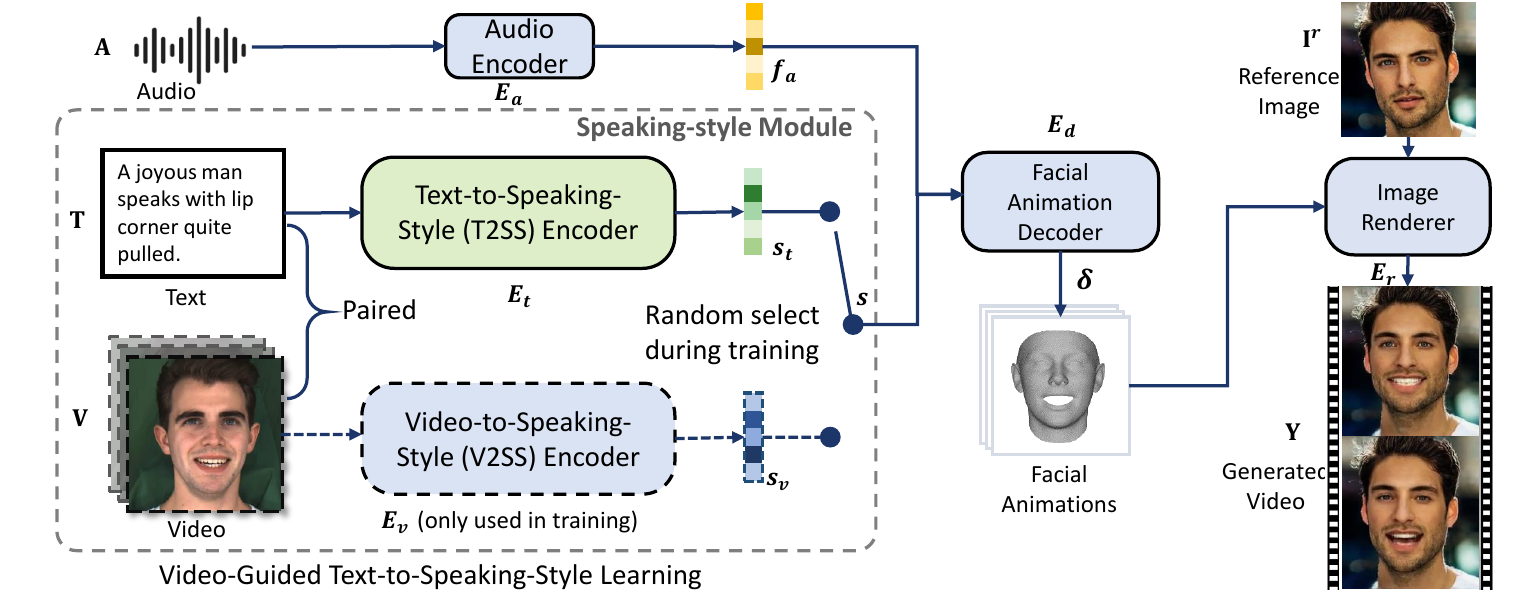}
  
  \caption{TalkCLIP pipeline. The text-to-speaking-style (T2SS) encoder can use natural language text to predict speaking style. By integrating the T2SS encoder with other modules, TalkCLIP can generate talking heads with text-guided speaking styles. To increase the alignment between text and predicted styles, we introduced a video-to-speaking-style encoder, which predicts speaking styles from video, to guide the training of the T2SS encoder. 
  % During training, two types of style codes are extracted from the video and text description, respectively. $S_v$ is extracted by a video-to-speaking-style (V2SS) encoder $E_v$ and used to guide the learning of $S_t$. During inference, only text provides the  speaking style code. 
  % An audio encoder extracts audio features from input audio.Then a decoder takes the  style code and the audio features as input and generates output expression parameters. Finally, an image renderer takes the expression parameters and the identity reference image and produces the output video. 
  }
  \label{fig:pipeline}
\end{figure*}

Given a reference image $\mathbf{I}^r$ of the target speaker, an audio clip $\mathbf{A}$ lasting $N$ frames, and a sentence $\mathbf{T}$ specifying the target speaking style, our framework aims to generate a talking head videos $\mathbf{Y}=\{\hat{\mathbf{I}}_i\}_{i=1}^{N}$, in which the target speaker utters the speech content with the target speaking style.
Note that our method focuses on facial expressions and does not generate head pose. The head pose information can be derived from real videos.

Our proposed framework is illustrated in \cref{fig:pipeline}. An audio encoder $E_a$ encodes the input audio into audio embeddings $f_a$. A text-to-speaking style (T2SS) encoder $E_t$ uses input text to generate the corresponding speaking style code $s$. Then, a facial animation decoder $E_d$ converts $f_a$ and speaking style code $s$ into expression animation parameters. A renderer $E_r$ takes the parameters as input and generates face frames. During training, we employ a video-to-speaking style (V2SS) encoder, which extracts style codes from videos, to guide the training of the T2SS encoder.

\subsection{Mapping Text to Speaking Styles}\label{text_to_speaking_style}
% In this paper, we aim to learn the semantic mapping from natural sentence description to facial motion patterns and control the speaking styles with text in an easy-to-understand manner. Apart from the direct control of speaking styles, we also explore inferring a speaking style from descriptions that may lie outside the distribution of our text annotations. It may require a bulk of paired visual-linguistic data for the text encoder to achieve this capacity. However, although we make efforts to enrich the text annotations, the textural descriptions in TA-MEAD are quite insufficient.

\textbf{CLIP-Based T2SS Encoder.} We aim to model the mapping from diverse natural language text to speaking styles. Since extensive knowledge of the natural language is stored in CLIP~\cite{radford2021learning}, we use CLIP as the backbone of T2SS encoder to facilitate generalization across real-world texts.
Since CLIP is trained with static text-image pairs while speaking styles depict the dynamic patterns of facial movements, it might be unsuitable to directly take the CLIP text embeddings as the speaking style codes. To adapt the CLIP text embeddings to the speaking style space, we design a lightweight adapter network. This adapter network follows the CLIP text encoder and tunes the clip-based text embedding $\mathbf{e}_t$ with additional residual connections to obtain the speaking-style code $\mathbf{s}_t$, which is formulated as follows:
\begin{equation}
    \mathbf{s}_t = \beta \mathbf{e}_s + (1 - \beta) \mathbf{e}_t ,
\end{equation}
where $\beta$ is used to control the tuning ratio between CLIP-based semantic features $\mathbf{e}_t$ and the adapter network output $\mathbf{e}_s$, and $\beta$ is set to 0.6 empirically. The residual connections preserve the prior semantic knowledge in CLIP. During training, the pre-trained CLIP text encoder is frozen, and only the adapter is optimized.

\noindent\textbf{Video-Guided T2SS Learning.}
It is challenging for the T2SS encoder to map natural language, a textual modality, to speaking styles, a visual modality. To ease this learning difficulty, we use a video-to-speaking-style (V2SS) encoder $E_v$, which can predict speaking styles from video, to guide the training of the T2SS encoder. The training process comprises two stages. First, together with other modules, the V2SS encoder is trained to predict style code from video. After the V2SS encoder can predict accurate speaking styles and hence can be a "good teacher", the training of the T2SS encoder starts.
As shown in Fig~\ref{fig:pipeline}, for each iteration, either the T2SS encoder or the V2SS encoder is randomly chosen to predict the style code for training. The probability that the T2SS encoder is chosen gradually increases from $0$ to $0.5$ as the training progresses. Besides, since we have constructed text-video pairs in TA-MEAD, we use a loss to align the style code derived from text $\mathbf{s}_t$ and the one derived from paired video. In this way, the T2SS can more easily learn the mapping from text to speaking style. During inference, only the T2SS encoder is used to infer speaking styles.

\subsection{Talking Head Video Generation}\label{talking_video_gen}
Leveraging the T2SS encoder that predict speaking styles from text, TalkCLIP can generate audio-driven talking head with text-guided speaking styles. To extract audio features, we use an automatic speech recognition tool to convert audio into a sequence of phoneme labels and then feed the labels into a transformer encoder to obtain audio embeddings $f_a$. 
Then, a face animation decoder $E_d$, which is mainly based on a transformer decoder, takes audio embeddings $f_a$ and predicted style code $s$ as input and produce a sequence of facial animations $\{\delta_i\}^N_{i-1}$. 
The face animation decoder predicts in a frame-by-frame manner. The decoder first constructs an audio embedding window for the current frame by taking several frames before and after the current frame. The window serves as the key in the transformer. The style code is repeated so that it has the same length as the audio window and then added with positional embeddings and serves the query. After obtaining the face animations, an image renderer PIRenderer~\cite{ren2021pirenderer} takes the animations and a portrait as input and generates the final talking videos.

\subsection{Training}
The end-to-end training stage optimizes the parameters of $E_a$, the adapter network in $E_t$,  $E_v$, and  $E_d$. $E_r$ is not included in the training stage. During training, we generate successive $L = 64$ frames of facial expression parameters $\mathbf{\delta}_{1:L}$ use a self-reconstruction setting. We design a reconstruction loss by combining L1 loss and structural similarity (SSIM) loss~\cite{wang2004image}:
\begin{equation}
    \mathcal{L}_\text{rec} = \mu \mathcal{L}_{
    \text{L1}}(\mathbf{\delta}_{1:L},\mathbf{\hat{\delta}}_{1:L}) + (1-\mu) \mathcal{L}_{\text{ssim}}(\mathbf{\delta}_{1:L},\mathbf{\hat{\delta}}_{1:L}),
\end{equation}
where $\mathbf{\delta}_{1:L}$ and $\mathbf{\hat{\delta}}_{1:L}$ are the ground truth sequence and reconstructed facial expression sequence, respectively. $\mu$ is a ratio coefficient and is set to 0.1. To stabilize the training process of the T2SS encoder, we use a cosine similarity loss to ensure the closeness of $\mathbf{s}_t$ and $\mathbf{s}_v$, extracted from the same text-video pair, throughout the training process:
\begin{equation}
\mathcal{L}_{\text {cos }} = 1 - \text{cos}(\mathbf{s}_t ,  \mathbf{s}_v).
\end{equation}
Additionally, we use an adversarial hinge loss $\mathcal{L}_\text{tem}$ to enhance temporal stability and a Lip-Sync Loss $\mathcal{L}_{\text{sync}}$ to improve lip synchronization.

% \noindent\textbf{Temporal Adversarial Loss.} We develop a temporal discriminator $D_\text{tem}$, which follows the PatchGAN structure~\cite{goodfellow2014generative,isola2017image,yu2017face,yu2017hallucinating,yu2018face}, to distinguish whether the generated sequence of expression parameters is real or not. By simultaneously training our framework with $D_\text{tem}$, we employ an adversarial hinge loss, denoted as $\mathcal{L}_\text{tem}$.

% \noindent\textbf{Lip-Sync Loss.} In order to enhance the accuracy of lip-sync, we develop a lip-sync discriminator $D_\text{sync}$ to calculate the probability of synchronization between the generated expression parameter sequence and the audio input. Based on the pretrained $D_\text{sync}$, we use a sync loss, denoted as $\mathcal{L}_{\text{sync}}$, to maximize the synchronous probability between $\mathbf{\delta}_{1:L}$ and the audio input.

% \noindent\textbf{Imitation Loss.} As mentioned before, to stabilize the training process of the T2SS encoder, we enforce $\mathbf{s}_t$ and $\mathbf{s}_v$ extracted from the same text-video pair to be close by using a cosine similarity loss during the whole training process:

% \begin{equation}
% \mathcal{L}_{\text {cos }} = 1 - \text{cos}(\mathbf{s}_t ,  \mathbf{s}_v).
% \end{equation}

Our total loss is a combination of the aforementioned loss terms:
\begin{equation}
\begin{split}
    \mathcal{L}=\lambda_{\text {rec }} \mathcal{L}_{\text {rec }}+\lambda_{\text {cos }} \mathcal{L}_{\text {cos }}+\lambda_{\text {sync }} \mathcal{L}_{\text {sync }}
    +\lambda_{\text {tem }} \mathcal{L}_{\text {tem }},
\end{split} 
\end{equation}
where we use $\lambda_{\text {rec }}=88$, $\lambda_{\text {cos }} = 1$, $\lambda_{\text {sync }} = 1$ and $\lambda_{\text {tem }} = 1$.

\section{Experimental Results}

\noindent\textbf{Datasets.} Our framework is trained on the proposed TA-MEAD dataset. We randomly select 42 speakers from the MEAD dataset and set aside 6 speakers for testing. Additionally, we evaluate our framework on the HDTF\cite{zhang2021flow} and Voxceleb2\cite{chung2018voxceleb2} datasets. For testing, we randomly select videos of 25 speakers from HDTF and follow the official testing settings of Voxceleb2.

\noindent\textbf{Metrics.}
% We conduct qantitative evaluations on
Several widely used metrics are employed to validate our method. The lip-sync is evaluated with the confidence score of SyncNet~\cite{chung2016out} ($\textbf{Sync}_{\textbf{conf}}$) and the Landmark Distance around mouths (\textbf{M-LMD})~\cite{chen2019hierarchical}. The facial expressions are evaluated with the Landmark Distance on the whole face (\textbf{F-LMD}). The video quality is evaluated with \textbf{SSIM} and (\textbf{CPBD})~\cite{narvekar2009no}.

% \begin{figure}
% \centering
% \includegraphics[width=0.98\linewidth]{images/au_int.pdf}
% \caption{(a) AU intensity control. (b) Emotion intensity control.}
% \label{fig:au_int}
% \end{figure}

\subsection{Quantitative results}

\begin{table*}[t]
    % \normalsize
    
    % \vspace{-2mm}
    \centering
    \caption{Quantitative comparisons are conducted on the MEAD, HDTF, and Voxceleb2 datasets. (GC-AVT is not included in the Voxceleb2 evaluation due to the unavailability of its samples on this dataset.) Although TalkCLIP merely derives speaking styles from text, which provide style information that is less detailed than videos, TalkCLIP still surpasses EAMM, GC-AVT, and PD-FGC in performance and rivals that of StyleTalk.}
    \resizebox{\textwidth}{!}{%
    \begin{tabular}{l|ccccccc}
    \toprule

    \multirow{2}{*}{Methods} & \multicolumn{5}{c}{MEAD / HDTF / Voxceleb2}   \\

        & SSIM$\uparrow$ & CPBD$\uparrow$ & F-LMD$\downarrow$ & M-LMD$\downarrow$ & $\text{Sync}_{\text{conf}}$ $\uparrow$   \\ 
        \midrule
        MakeItTalk~\cite{zhou2020makelttalk} &        0.73 / 0.57 / 0.52 & 0.11 / 0.20 / 0.24 & 3.95 / 5.12 / 6.29 & 5.39 / 4.61 / 5.15 & 2.15 / 3.20 / 2.17  \\ 
        Wav2Lip~\cite{prajwal2020lip} &        0.81 / 0.59 / 0.54 & \textbf{0.16} / \textbf{0.26} / \textbf{0.30} & 2.73 / 5.11 / 5.85 & 3.85 / 3.84 / 4.64 & \textbf{5.41} / 4.57 / \textbf{5.70}  \\ 
        PC-AVS~\cite{zhou2021pose} &        0.51 / 0.42 / 0.36 & 0.07 / 0.12 / 0.09 & 5.87 / 10.7 / 12.9 & 5.03 / 4.26 / 7.42 & 2.21 / 4.15 / \underline{4.73}  \\ 
        AVCT~\cite{wang2022one} &        \underline{0.83} / 0.74 / 0.64 & 0.14 / 0.18 / 0.23 & 2.95 / 3.06 / 3.62 & 5.64 / 3.83 / 3.71 & 2.56 / 4.46 / 3.89  \\ 
        GC-AVT~\cite{liang2022expressive} &        0.34 / 0.33 / \ \, - \ \,\, & 0.14 / 0.24 / \ \, - \ \,\, & 8.11 / 10.7 / \ \, - \ \,\, & 8.43 / 6.34 / \ \, - \ \,\, & 2.41 / 4.23 / \ \, - \ \,\, \\ 
        EAMM~\cite{ji2022eamm} &        0.40 / 0.37 / 0.43 & 0.08 /  0.13 /  0.20 & 6.67 /  7.74 /  6.36 & 6.60 /  7.67 /  4.89 & 1.42 /  2.78 /  2.24  \\ 
        SadTalker~\cite{zhang2023sadtalker} &        0.68 /  0.73 /  0.44 & \textbf{0.16} /  0.19 /  0.19 & 4.04 /  6.26 /  9.12 & 4.24 /  4.18 /  6.11 & 2.87 /  3.86 /  4.38  \\ 
        PD-FGC~\cite{wang2023progressive} &        0.51 /  0.40 /  0.35 & 0.05 /  0.13 /  0.12 & 5.41 /  9.99 /  12.5 & 3.94 /  4.46 /  8.19 & 2.46 /  4.20 /  4.64  \\ 
        EAT~\cite{gan2023efficient} &        0.53 /  0.55 /  0.47 & 0.15 /  0.18 /  0.20 & 5.63 /  4.12 /  5.53 & 4.98 /  4.24 /  5.88 & 2.19 /  3.95 /  4.35  \\ 
        StyleTalk~\cite{ma2023styletalk} &        \textbf{0.84} /  \textbf{0.80} /  \underline{0.66} & \textbf{0.16} /  \textbf{0.26} /  \underline{0.29} & \textbf{2.17} /  \textbf{2.04} /  \textbf{2.92} & \textbf{3.36} /  \textbf{2.50} /  \textbf{2.96} & 3.51 /  \textbf{4.75} /  4.51  \\ 
        Ground Truth &        1 /  1 /  1 & 0.22 /  0.26 /  0.33 & 0 /  0 /  0 & 0 /  0 /  0 & 4.18 /  5.27 /  5.23  \\ 
        \textbf{TalkCLIP} &        \underline{0.83} /  \underline{0.78} /  \textbf{0.67} & \textbf{0.16} /  \underline{0.25} /  \underline{0.29} & \underline{2.42} /  \underline{2.54} /  \underline{2.94} & \underline{3.60} /  \underline{2.84} /  \underline{2.99} & \underline{3.77} /  \underline{4.69} /  4.60 \\

    \bottomrule
    \end{tabular}
    } %

    \label{table:quantitive_evaluation}
\end{table*}

% \textit{\textbf{Baselines.}}
Our method is compared with state-of-the-art methods including: MakeitTalk~\cite{zhou2020makelttalk}, Wav2Lip~\cite{prajwal2020lip}, PC-AVS~\cite{zhou2021pose}, AVCT~\cite{wang2022one}, GC-AVT~\cite{liang2022expressive},  EAMM~\cite{ji2022eamm}, StyleTalk~\cite{ma2023styletalk},  SadTalker~\cite{zhang2023sadtalker}, PD-FGC~\cite{wang2023progressive}, and EAT~\cite{gan2023efficient}. The experiments are conducted in a self-driven setting on the test set. We select the first image of each video as the reference image, and the corresponding audio clip as the audio input. GC-AVT, EAMM, StyleTalk, and PD-FGC take the GT (ground truth) video as the speaking style reference. For MEAD, our method extracts the speaking styles from the text annotations of the GT videos. Since most speakers exhibit neutral expressions in HDTF and Voxceleb2, we use the textual description "An expressionless woman/man" as the speaking style reference. The samples of the compared methods are generated either with their released codes or with authors' help.
% Since Wav2Lip only generates movements of the mouth area, the head poses are fixed in its samples. For other methods, poses are derived from ground truth videos. (COPY FROM STYLETALK)

% We also compare our method with speaker-specific emotional talking head methods \textbf{Write-a-speaker} \cite{li2021write} and \textbf{EVP} \cite{ji2021audio} and the results are reported in the supplementary material.

% \begin{table*}
% \centering
% \footnotesize

% % \vspace{-1em}
% \setlength{\tabcolsep}{0.6mm}{\begin{tabular}{ccccccccccccc}
% \toprule  
% Method & Wav2Lip & MakeitTalk & PC-AVS & AVCT & EAMM & GC-AVT &SadTalker & PD-FGC & EAT & StyleTalk & Ground Truth & TalkCLIP \\
% \midrule
% Lip Sync & 3.63 & 2.19 & 3.47 & 3.56 & 1.92 & 3.34 & & & & 3.61 & 4.45  & \textbf{3.70} \\
% Video Realness & 1.68 & 2.31 & 1.96 & 2.82 & 2.05 & 1.43 & & & & \textbf{3.07} & 4.29  & \underline{3.04} \\
% Style Consistency with Video & - & -  & - & - & 2.35 & 2.84 & - & &  & \textbf{3.77} & -  & \underline{3.53} \\
% Style Consistency with Text & - & -  & - & - & 2.09 & 2.41 & - & & &  3.44  & -  & \textbf{3.65} \\
% \bottomrule 
% \end{tabular}
% }
% \caption{Results of the user study for video quality. }
% \label{table:user_study}

% \end{table*}

As presented in Table~\ref{table:quantitive_evaluation}, our method performs better than all other methods, except StyleTalk, across most metrics. It is worth noting that our method solely infers speaking styles from easily accessible text descriptions, which is a more challenging task compared to EAMM, GC-AVT, StyleTalk, and PD-FGC that extract speaking style information from videos. As previous expression-controllable methods extract speaking styles from the corresponding GT videos and thus obtain more concrete information than our method from texts, EAMM, GC-AVT, StyleTalk, and PD-FGC have an advantage in most metrics. Despite this, our method still outperforms EAMM, GC-AVT, and PD-FGC and achieves competitive performance compared to StyleTalk.

% As shown in Table~\ref{table:quantitive_evaluation}, Our method outperforms all other methods except StyleTalk in most metrics. Note that our method merely infers speaking style information from easy-to-use text descriptions, which is a more challenging task compared to  EAMM, GC-AVT, and StyleTalk which extract speaking style information from videos.
% Since previous expression-controllable methods extract speaking styles from the corresponding GT videos and videos provide more concrete information than texts, EAMM, GC-AVT and StyleTalk have advantages in most metrics. Despite the advantages of these methods, our method still outperforms EAMM and GC-AVT and achieves competitive performance compared to StyleTalk. 
% Since videos provide more concrete and accurate information on speaking style than texts, EAMM, GC-AVT, and StyleTalk 

\begin{figure*}[t]
  \centering
  \includegraphics[width=0.98\textwidth]{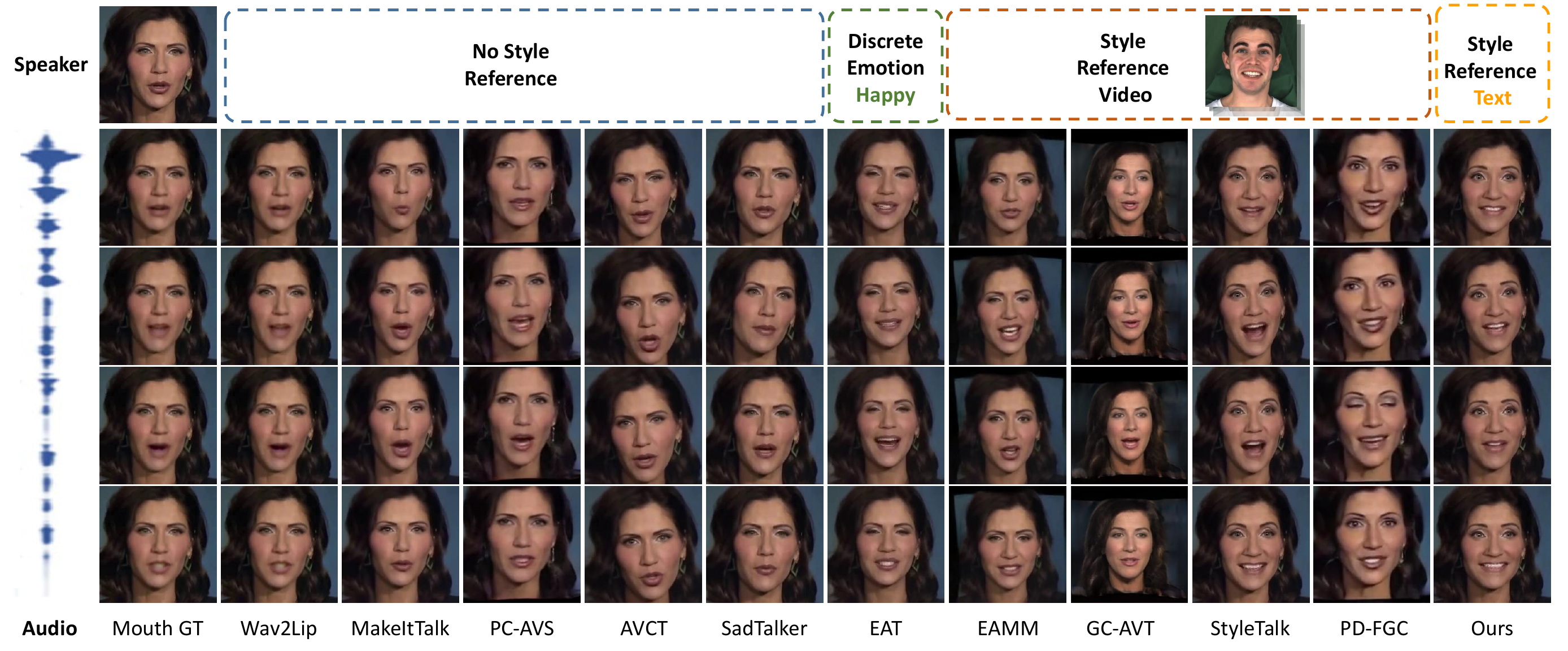}
  \caption{Qualitative comparisons. Note that the speaking style of our method is derived from text description. The style reference text is \emph{A woman feels ecstatic and speaks with fairly lifted cheek, strongly raised outer brow, and lip corner fully pulled}.}
  % Style reference text B is \emph{A woman feels ecstatic and speaks with fairly lifted cheek, strongly raised outer brow, and lip corner fully pulled}.  Please zoom in or refer to our demo video for more details.}
  \label{fig:qualitative}
\end{figure*}

It is observed that PC-AVS and PD-FGC achieve higher SyncNet confidence scores on Voxceleb2, likely due to their training on this specific dataset, which may provide advantages in comparisons. However, they demonstrate poor LMD scores and tend to yield blurry results.
We also note that Wav2Lip obtains a higher confidence score of SyncNet on MEAD and Voxceleb2 than our method, and the score is even higher than that of ground truth. However, since SyncNet serves as a discriminator when Wav2Lip is trained, it is reasonable for Wav2Lip to obtain a high score in this metric. Note that our M-LMD score is better than that of Wav2Lip and close to that of GT. These results indicate that our method generates realistic talking heads with proper speaking styles guided by text while achieving accurate lip-sync compared with the state-of-the-art.
% This shows that our method is able to attain accurate lip-sync.

\subsection{Qualitative results}

Figure~\ref{fig:qualitative} shows the qualitative results. The identity reference and audio are unseen during training. Among the methods, only EAMM, GC-AVT, StyleTalk, PD-FGC, EAT and our method can control speaking styles. EAMM, GC-AVT, StyleTalk, and PD-FGC require an additional video as the style reference, while our method solely relies on a text description, making it more convenient. EAMM and GC-AVT can only control the upper facial expressions, they fail to control the speaking styles in the mouth area. EAMM also fails to achieve accurate lip-sync and natural head pose changes. GC-AVT and PD-FGC fail to preserve the speaker's identity. EAT can only generate coarse-grained emotions and cannot control personalized idiosyncrasies. The speaker's eyes appear closed in the EAT's samples, contrasting with the open eyes observed in the reference video. Our method and StyleTalk control the speaking styles across the entire face, achieve accurate lip-sync, preserve identity, and generate plausible backgrounds. However, StyleTalk requires a video that expresses the desired speaking style, whereas our method only requires a simple text description. Consequently, our method generates expressive talking heads by controlling the speaking style in a more convenient manner.

% Only EAMM, GC-AVT, StyleTalk and our method are able to control speaking styles. EAMM, GC-AVT, and StyleTalk need to take an additional video as the style reference, while our method merely uses the text description, which is much more convenient. EAMM and GC-AVT can only control the upper facial expressions but fail to control the speaking styles in the mouth area. EAMM fails to achieve accurate lip-sync and the head pose change is unnatural. GC-AVT fails to preserve the speaker's identity and cannot generate plausible backgrounds. Our method and StyleTalk both control the speaking styles in the entire face while achieving accurate lip-sync, satisfactory identity preservation and generating plausible backgrounds. However, StyleTalk needs a video that expresses the desired speaking style, while our method merely needs a straightforward text description. Therefore, our method can generate expressive talking heads by controlling the speaking style in a more convenient manner. Please refer to our demo video for more details.

\subsection{Ablation Study}

\begin{figure}[t]
    \centering\small
    \begin{minipage}[t!]{0.48\textwidth}
        \centering\small
        \includegraphics[width=\columnwidth]{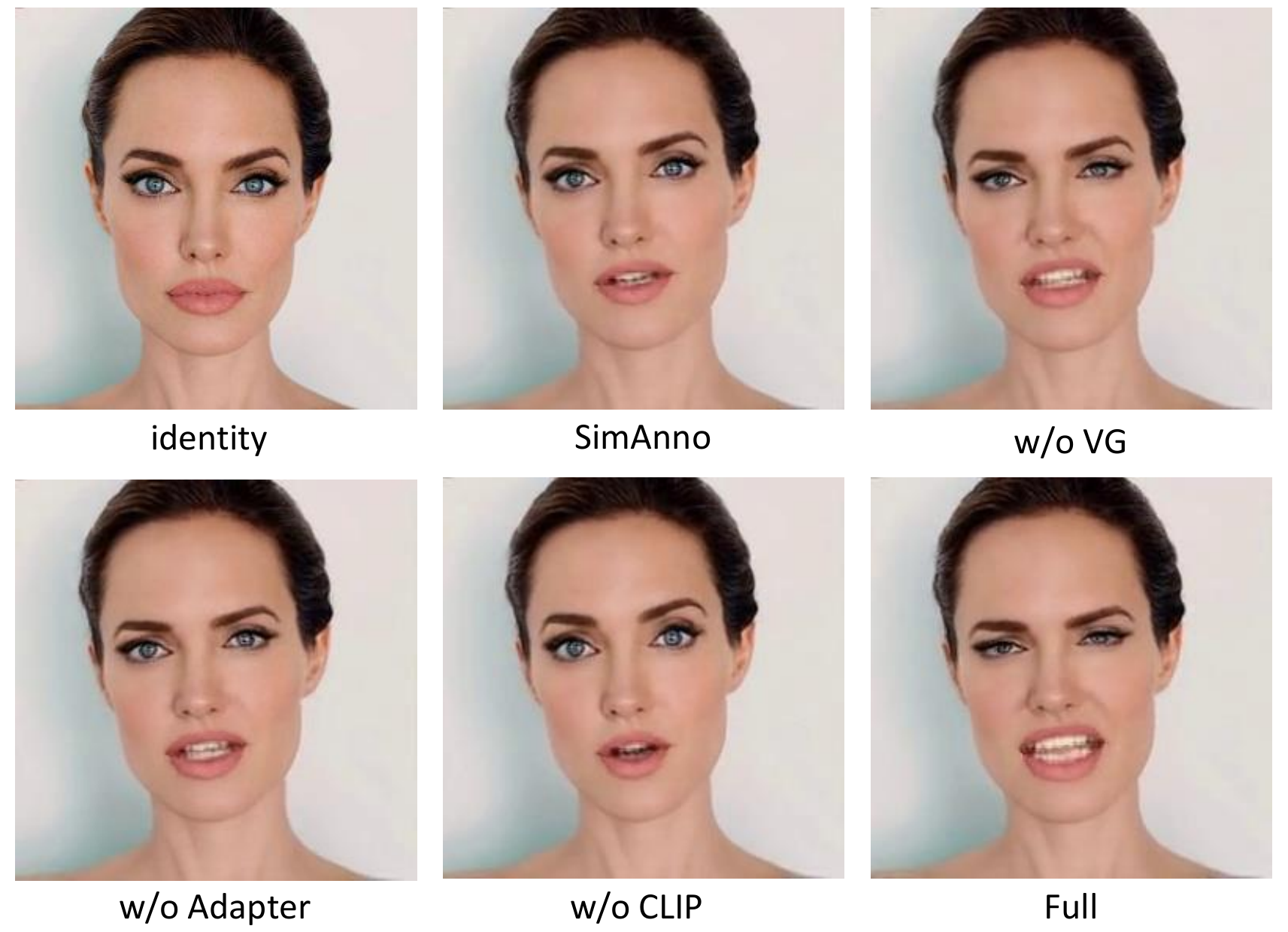}
        \caption{%
        Qualitative results of the ablation study. The speaking style is inferred from the out-of-domain text "A woman screwed up her face".
        }
        \label{fig:ablation}
    \end{minipage}
    \hskip1em
    \begin{minipage}[t!]{0.48\textwidth}
        \centering\small
        \includegraphics[width=\columnwidth]{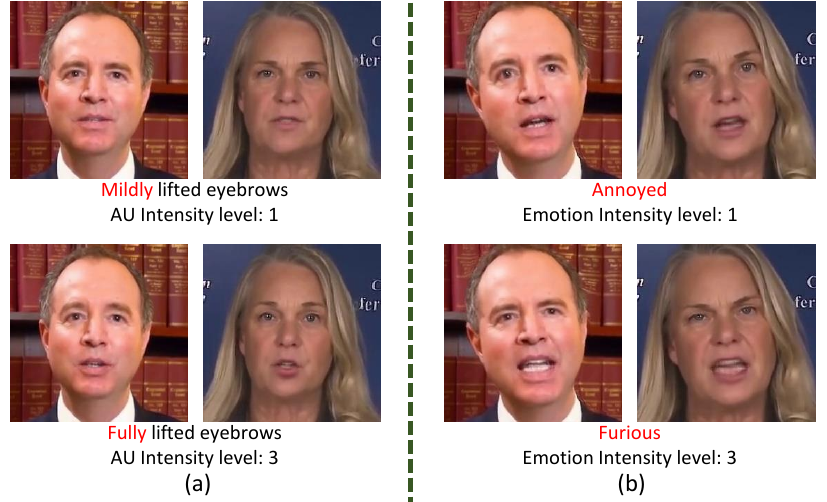}
        \caption{%
        (a) AU intensity control. (b) Emotion intensity control.
        }
        \label{fig:au_int}
    \end{minipage}
\end{figure}

\begin{table}
\footnotesize
\centering
\caption{Quantitative results of the ablation study on MEAD.}
\setlength{\tabcolsep}{1.3mm}{
\begin{tabular}{cccccc}
\toprule  
Method & SSIM$\uparrow$ &  CPBD$\uparrow$ & F-LMD $\downarrow$  & M-LMD $\downarrow$ & $\text{Sync}_{conf}$$\uparrow$ \\
\midrule

SimAnno     & 0.826 &  \textbf{0.164} & 2.858 & 3.815 & 3.756 \\
w/o Adapter      & \textbf{0.829}  & 0.162 & 2.761 & 3.739 & 3.735 \\
w/o VG  & 0.827  & 0.163 & 2.74 & 3.704 & 3.752 \\
w/o CLIP    & \textbf{0.829}  & 0.162 & \textbf{2.397}   & \textbf{3.593} & \textbf{3.791}  \\
\cmidrule(r){1-6}
Full    & \textbf{0.829}  & \textbf{0.164} & \underline{2.415}  & \underline{3.601} & \underline{3.773} \\
\bottomrule 
\end{tabular}}

\label{table:ablation_study}
\end{table}

To evaluate the effectiveness of our design, we perform ablation studies with 5 variants: (1)\textbf{SimAnno}: only use the emotion labels (eg. angry, happy)  to annotate videos, (2)\textbf{w/o Adapter}: remove the adapter network in $E_t$, (3) \textbf{w/o CLIP}: avoid using the pre-trained weights of CLIP, (4) \textbf{w/o VG}: train models without video-guided learning, (5) \textbf{Full}: our full model. 

The quantitative results on MEAD are reported in Table~\ref{table:ablation_study} and the qualitative results are shown in Figure~\ref{fig:ablation}. Note that the speaking style of the results in Figure~\ref{fig:ablation} is guided by the out-of-domain text description.
Since all variants use the same image renderer and lip-sync discriminator, they attain similar SSIM and CPBD and $\text{Sync}_{conf}$ scores. Compared with Full, the lower F-LMD and M-LMD scores of SimAnno imply that simple emotion labels are insufficient to learn the semantic mapping from natural language to speaking styles, validating the effectiveness of the proposed annotation approach.
Based on the results of w/o Adapter and w/o VG in Table~\ref{table:ablation_study} and Figure~\ref{fig:ablation}, we can find that both the adapter network and the video-guided training contribute to aligning the clip embeddings to the speaking style space. Although w/o CLIP achieves competitive quantitative results on the MEAD test set using the annotation text compared to Full, our qualitative results indicate that pre-trained CLIP significantly improves the generalization of our approach to out-of-domain text.

\subsection{User Study}

\begin{table*}[t]
\centering
\footnotesize
\caption{Results of the user study for video quality. LS, VR, SCV, SCT represents Lip Sync, Video Realness, Style Consistency with Video/Text, respectively.}
% \vspace{-1em}
\resizebox{\textwidth}{!}{\begin{tabular}{ccccccccccccc}
\toprule  
 & Wav2Lip & MakeitTalk & PC-AVS & AVCT & EAMM & GCAVT &SadTalker & PD-FGC & EAT & StyleTalk & GT & TalkCLIP \\
\midrule
LS & 3.42 & 1.98 & 3.34 & 3.21 & 1.72 & 3.37 & 3.47 & 3.29 & 3.11 & 3.53 & 4.19  & \textbf{3.76} \\
VR & 1.46 & 2.12 & 1.79 & 2.70 & 1.51 & 1.69 & 2.84 & 1.98 & 2.66 & \textbf{3.01} & 3.84  & \underline{3.00} \\
SCV & - & -  & - & - & 2.01 & 2.28 & - & 2.55 & 2.43 & \textbf{3.54} & -  & \underline{3.27} \\
SCT & - & -  & - & - & 1.62 & 1.83 & - & 2.28 & 2.36 &  3.21  & -  & \textbf{3.51} \\
\bottomrule 
\end{tabular}
}
\label{table:user_study}
\end{table*}

We conduct user studies of 34 participants to evaluate the quality of TalkCLIP's results, the quality of TA-MEAD text descriptions, and TalkCLIP's generalization capability for open-ended text prompts.

\begin{figure}[t]
\centering
\includegraphics[width=0.6\linewidth]{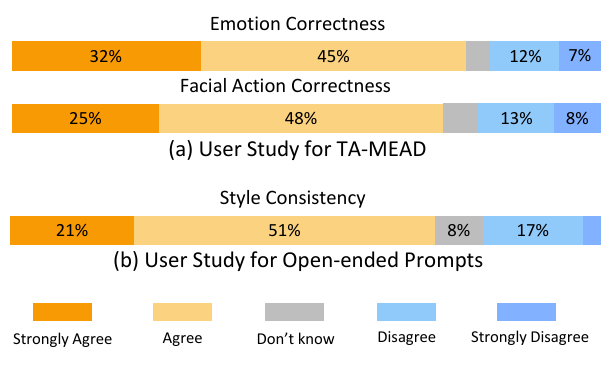}
\caption{User study results for TA-MEAD (a) and open-ended prompts (b)}
\label{fig:userstudy}
\end{figure}

\noindent\textbf{Evaluating the quality of generated videos} Our questionnaire includes five videos for each method on the MEAD test dataset. For emotion-controllable methods (EAMM, GC-AVT, PD-FGC, EAT, StyleTalk, and TalkCLIP), we include style reference videos and corresponding text descriptions in the questionnaire. Participants rate the videos on a scale of 1 to 5 on four aspects: (1) lip synchronization; (2) video realness; (3) style consistency with the reference video; and (4) style consistency with the reference text.

Table~\ref{table:user_study} shows the results. We exclude the scores of style consistency for style-uncontrollable methods. Our method achieves the best lip sync score. Since both StyleTalk and TalkCLIP employ PIRenderer as the renderer, their video realness scores are close and are higher than those of other methods. Despite our method's use of text as the style reference, our scores for \emph{style consistency with video} are better than EAMM and GC-AVT, PD-FGC, and competitive with StyleTalk. Our method also outperforms all others in the \emph{style consistency with text} score. The improved results over EAT illustrate the advantages of guiding speaking styles with natural language, rather than using emotion labels.

\noindent\textbf{Evaluating the quality of TA-MEAD Text Descriptions} We conduct an evaluation of the quality of our prompt-based text descriptions in TA-MEAD. Our questionnaire includes 20 text-video pairs sampled from TA-MEAD for each participant. Participants are asked to express whether they agree with two statements: \emph{(1) Emotion Correctness: whether the emotion part in the text accurately describes the emotions in the video}, and \emph{(2) Facial Action Correctness: whether the facial action part in the text accurately describes the facial actions in the video}.

As shown in Figure~\ref{fig:userstudy}, 77\% and 73\% of the text descriptions accurately describe those emotions and facial actions, respectively. This validates the effectiveness of our prompt-based annotations. Moreover, we also inspect samples with low scores in the MEAD dataset. We observed that some videos are categorized with erroneous emotion labels in the original MEAD dataset (e.g., some neutral emotion videos are placed in the "surprise" category), which lowers the emotion correctness. We also observe that the AU detector generates erroneous AU intensities for a small number of videos, lowering the facial action correctness. We manually correct the annotations that participants in the user study consider incorrect.

\noindent\textbf{Evaluating generalization capability for open-ended prompts} 
To investigate the generalizability of TalkCLIP to open-ended text inputs, we request participants to create their own prompts and evaluate the generated videos. Participants are instructed to create prompts that include at least one prompt for each of the 7 emotions (e.g. happy, angry, etc.). The generated prompts are used to create videos, and participants are asked to express whether they agree that the speaking style in generated videos are consistent with the prompts. Figure \ref{fig:userstudy} shows that 72\% of the samples are considered to be style-consistent, indicating that TalkCLIP is able to generalize to open-ended prompts with satisfactory style consistency.

\subsection{More results and Analysis}

\begin{figure}[t]
    \centering\small
    \begin{minipage}[t!]{0.48\textwidth}
        \centering\small
        \includegraphics[width=\columnwidth]{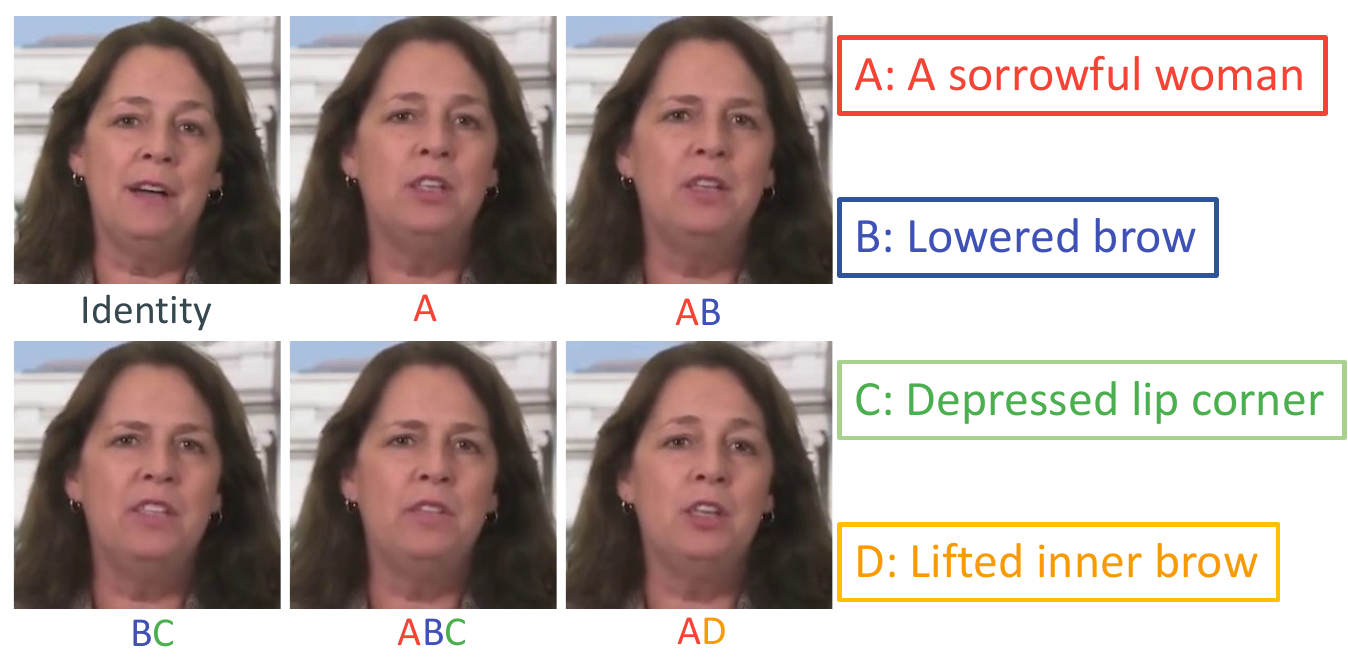}
        \caption{%
        Visualize the effect of each text part on the speaking styles.
        }
        \label{fig:contri}
    \end{minipage}
    \hskip1em
    \begin{minipage}[t!]{0.48\textwidth}
        \centering\small
        \includegraphics[width=\columnwidth]{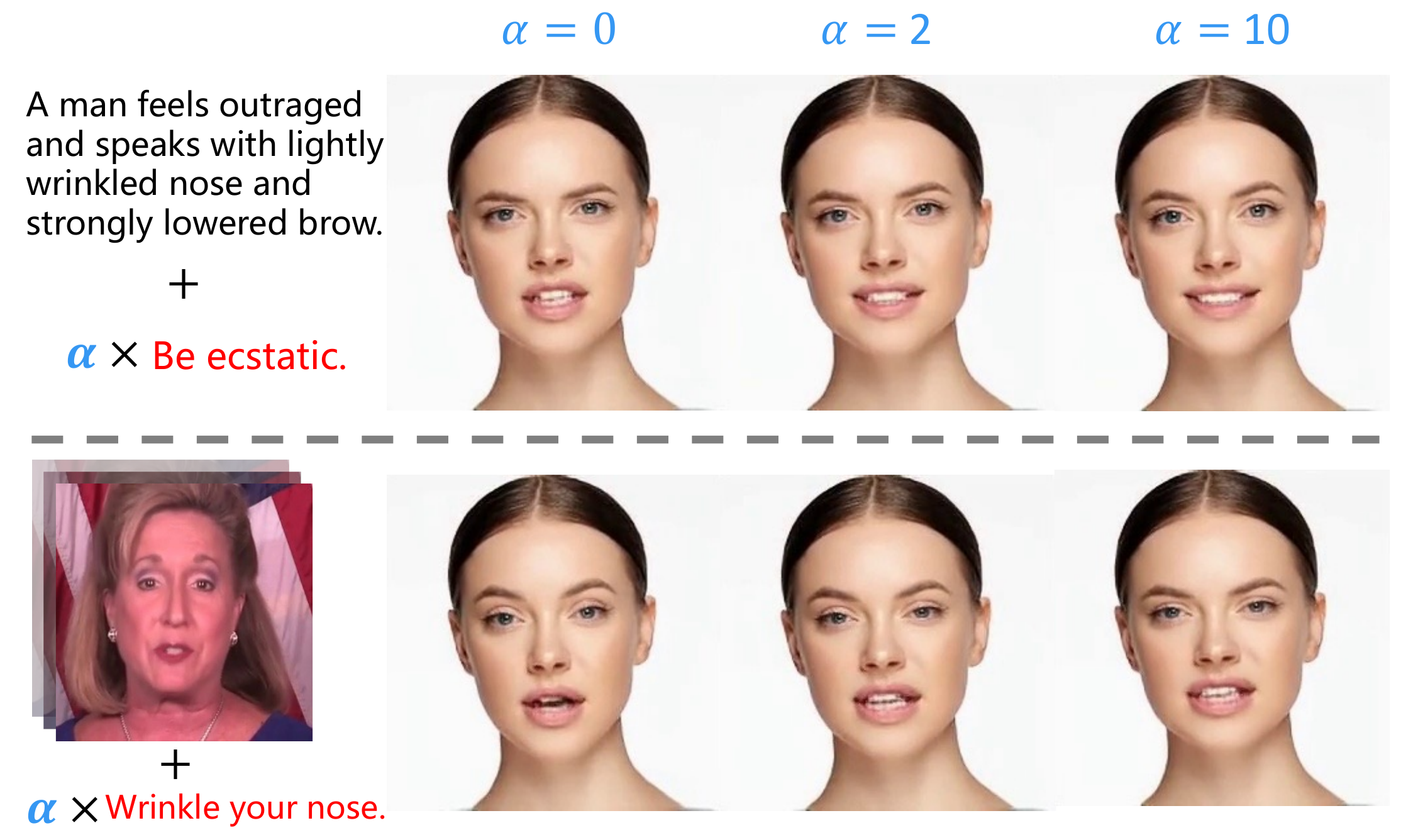}
        \caption{%
        TalkCLIP can edit speaking styles from texts (top) or videos (bottom) by adding style codes of texts that describe the desired edits. $\alpha$ controls the degree of editing. The speaker in the portrait can be different from the speaker in the style reference video (bottom).
        }
        \label{fig:editing}
    \end{minipage}
\end{figure}

\textbf{Generalizing to Out-of-Domain Text.}
As shown in Figure~\ref{fig:teaser}, our method is able to infer speaking styles from out-of-domain natural language. This capability is crucial for real-world applications. Notably, even though our text annotations do not include the word "eye", our method can still generate a "wide-eyed" speaking style. Similarly, our method can generate a "wince" speaking style even when the word "wince" is not present in our annotations. 
% These results demonstrate the effectiveness of our method in generalizing well to text descriptions that are outside of the annotated domain.
% This capacity is vital to real-world applications. Specifically, while our text annotations do not contain the word "eye", our method is able to generate a "wide-eyed" speaking style. Similarly, while the word "wince" does not exist in our annotations, our method is able to generate a "wince" speaking style. These results demonstrate that our method can generalize well to out-of-domain text descriptions. 

\noindent\textbf{Emotion and AU Intensity Control.}
We visualize the effects of AU intensity and emotion intensity by altering the adverbs or emotion labels with varying intensity levels. The experiments demonstrate our method's ability to perceive the differences in emotion implied in text descriptions with distinct intensity levels, and then translate them into corresponding facial expressions. As shown in Figure~\ref{fig:au_int}, our method generates a higher intensity speaking style for "Fully" compared to "Mildly," and also for "Furious" compared to "Annoyed." 

\noindent\textbf{Analysis on Text Prompt.}
We analyze the contribution of each text prompt to the inferred speaking styles. In Figure~\ref{fig:contri}, we divide emotion and AUs into several text description components and inspect the speaking style inferred from different combinations of these components. We observe that facial actions alone (the combination of BC in Figure~\ref{fig:contri}) can also generate the described speaking style, which is consistent with the fact that people sometimes use facial actions to convey expressions.
These results demonstrate that each component has an impact on the speaking styles, indicating the ability of our method to capture the fine-grained facial expressions implied in each text description. Additionally, we find that the order of prompts does not affect the generated facial expressions.

% We analyze the contribution of each text part to the inferred speaking styles. As shown in Figure~\ref{fig:contri}, we split emotion and AUs into several text description components and inspect the speaking style inferred from different combinations of these components. It shows that each component has an influence on the speaking styles.
% This phenomenon demonstrates that our method is able to capture the fine-grained facial expressions implied in each text description.
% We observed that using text that only describes facial actions (combination BC in Figure~\ref{fig:contri}) can also generate the described speaking style. This fits the fact that people sometimes also only use facial actions to describe expressions. We provide more analysis of the influence of texts on speaking styles in our \emph{supplementary materials}.
% Specifically, combination A generates a sad speaking style. When adding components B or D, the brows are lowered or lifted, respectively. When adding component C, the lip corner are depressed. 

% generate different fine-grained speaking styles for the same emotion by using text descriptions. 

\noindent\textbf{Text-Guided Speaking Style Editing.}
TalkCLIP can edit speaking styles by using additional text. The additional text can describe coarse-level emotion or fine-grained facial movements. Specifically, a style code is derived from the additional text and then added to the style code derived from the original input text with a weight that controls the degree of editing. The first row of \cref{fig:editing} shows an example where the speaking style is modified from angry to happy through this editing process.

Moreover, TalkCLIP can also edit the speaking style from videos. We can add the style code inferred from the additional text to style codes extracted from a video using the V2SS encoder. The bottom row of \cref{fig:editing} demonstrates an example where an additional text causes a speaker to wrinkle the nose. 
% This feature has numerous potential applications, such as expression editing in film post-production.

\section{Limitations}
Although our method has achieved promising results in text-guided speaking style control, it still has a few limitations that present novel research opportunities. Firstly, TalkCLIP struggles to generate correct speaking styles for texts that are too abstract, such as idioms (e.g., "He was over the moon about the ratings"). Secondly, expressions derived from texts tend to show lower intensity than those derived from videos, especially when the texts only describe coarse emotions and do not depict fine-grained facial movements. This is because texts are a more ambiguous signal than videos. A text may correspond to a group of videos. During training, TalkCLIP only learns the mean style of the group, hence reducing the expression intensity. Using texts that describe expressions in detail alleviates the issue. Thirdly, TalkCLIP may produce artifacts around the mouth when generating extreme expressions. The issue can be solved by using more advanced rendering techniques. Despite these limitations, we argue that TalkCLIP represents a significant step towards convenient text-guided speaking style control.

\section{Conclusion}
This paper introduces TalkCLIP, a system that generates expressive talking faces with text-guided speaking styles, and TA-MEAD, a talking head dataset with diverse fine-grained textual descriptions for speaking styles. At the core of TalkCLIP is a CLIP-based text encoder that can map real-world natural language texts to speaking styles. TalkCLIP can use text to modulate expression intensity and edit expressions, which is useful for applications such as CG facial animation editing. Extensive experiments validate the effectiveness of TalkCLIP and the quality of TA-MEAD. We hope that our work will inspire future research in text-to-face-motion generation.

\bibliographystyle{IEEEtran}
\bibliography{main}

\end{document}